\documentclass[10pt,twocolumn,letterpaper]{article}
\usepackage{iccv}
\usepackage{times}
\usepackage{epsfig}
\usepackage{graphicx}
\usepackage{amsmath}
\usepackage{amssymb}
\usepackage{authblk}
\usepackage{algorithm}
\usepackage{algorithmic}
\usepackage{subfigure}

\usepackage{bm}
 \usepackage{multirow}
\usepackage[table,xcdraw]{xcolor}
\newcommand{\ignore}[1]{}
\usepackage[pagebackref=true,breaklinks=true,letterpaper=true,colorlinks,bookmarks=false]{hyperref}

\definecolor{rowblue}{RGB}{220,230,240}
\definecolor{myorchid}{RGB}{150,10,30}
\definecolor{myblue}{RGB}{10,30,250}
\definecolor{mygreen}{RGB}{10,120,10}
\definecolor{grey}{RGB}{100,100,100}

\def\iccvPaperID{3624} 

\iccvfinalcopy 
\begin{document}

\title{Bounding Box Annotation for Visual Tracking via Selection and Refinement}
\title{Video Annotation for Visual Tracking via Selection and Refinement}
\author{Kenan Dai\textsuperscript{\dag}, Jie Zhao\textsuperscript{\dag}, Lijun Wang\textsuperscript{\dag}\thanks{Corresponding Author: Dr. Lijun Wang}, Dong Wang\textsuperscript{\dag}, Jianhua Li\textsuperscript{\dag}, 
Huchuan Lu\textsuperscript{\dag}, Xuesheng Qian\textsuperscript{\S}, Xiaoyun Yang{\ddag}\\
{\dag} Dalian University of Technology, China,
{\S} CSA Intellicloud Ltd, 
{\ddag} Remark Holdings \\
{\sl{\small{dkn10088@gmail.com, zj982853200@mail.dlut.edu.cn, xingkong19890806@gmail.com, \{wdice,jianhual,lhchuan\}@dlut.edu.cn, xuesheng.qian@intellicloud.ai, xyang@remarkholdings.com}}}
}

\maketitle
\ificcvfinal\thispagestyle{empty}\fi

\begin{abstract}
 Deep learning based visual trackers entail offline pre-training on large volumes of video datasets with accurate bounding box annotations that are labor-expensive to achieve. We present a new framework to facilitate bounding box annotations for video sequences, which investigates a selection-and-refinement strategy to automatically improve the preliminary annotations generated by tracking algorithms. A temporal assessment network (T-Assess Net) is proposed which is able to capture the temporal coherence of target locations and select reliable tracking results by measuring their quality. Meanwhile, a visual-geometry refinement network (VG-Refine Net) is also designed to further enhance the selected tracking results by considering both target appearance and temporal geometry constraints, allowing inaccurate tracking results to be corrected. The combination of the above two networks provides a principled approach to ensure the quality of automatic video annotation. Experiments on large scale tracking benchmarks demonstrate that our method can deliver highly accurate bounding box annotations and significantly reduce human labor by 94.0\%, yielding an effective means to further boost tracking performance with augmented training data. 
\end{abstract}

\section{Introduction}

\ignore{
1. why is this problem important

2. what are the major challenges

3. drawbacks of existing methods

4. how is your method different from others

5. why does your method make sense

6. what are your major contributions

7. how is your performance compared with others
}
Visual tracking aims to address the challenging problem of video target localization based on target appearance models.
Recent studies~\cite{SiamFC,dasiamrpn,StructSiam,siamrpn++,Ocean_2020_ECCV} propose to perform tracking with offline pre-trained deep features, yielding record-breaking results on most benchmarks. Their success is highly reliant on the availability of large-scale video datasets~\cite{huang2019got10k,real2017youtube,lasot,trackingnet} with accurate annotations. However, manually annotating target bounding boxes is tedious and labor-intensive. Therefore, labeled datasets for training visual trackers are still very rare and expensive to achieve, which restricts the potential performance boost of existing tracking algorithms.  

To mitigate the above issue, some recent works~\cite{trackingnet,vondrick2010efficiently,vondrick2011video,manen2017pathtrack} explore machine learning techniques to facilitate video annotation. The basic principle is to ask human annotators to label ground truth bounding boxes for only a sparse set of frames, while the reset annotations are automatically produced using either temporal interpolation or state-of-the-art tracking algorithms. 
Significant progress has been achieved by recent studies along this line which effectively reduce human labors required by video annotation.

\begin{figure}[t]
\begin{center}
\includegraphics [width=0.98\linewidth]{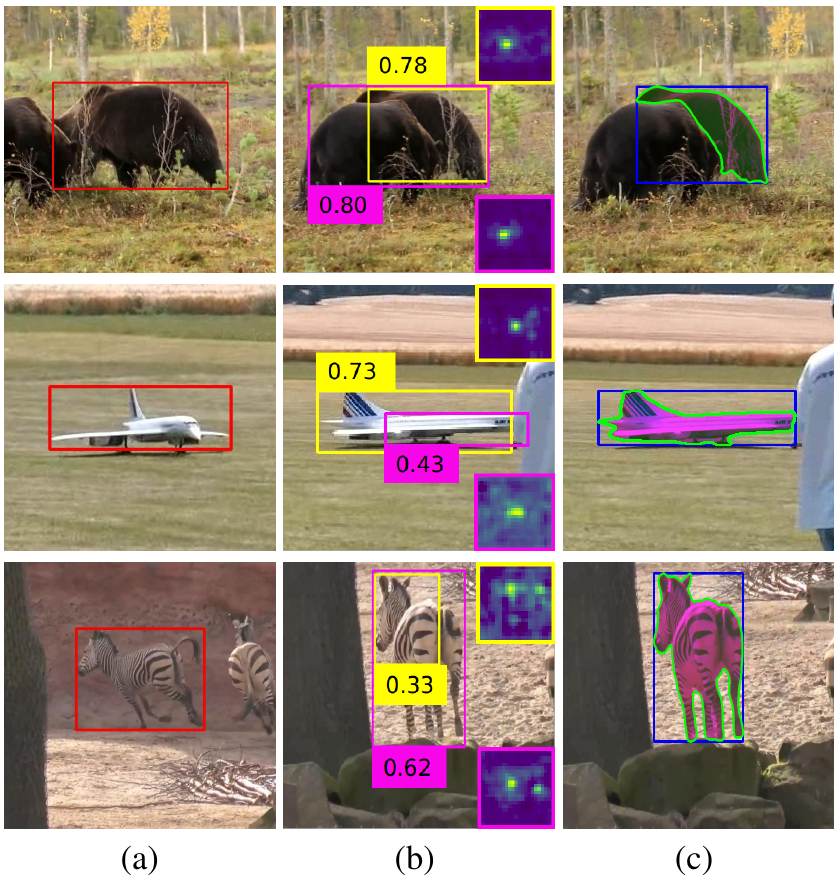}
\end{center}
   \caption{Visualization of our intermediate results. (a) Initial frame with manual annotations. (b) A subsequent frame with preliminary forward (yellow) and backward (pink) tracking results, and the predicted quality scores. (c) Results of target region inference and the generated annotations.}
\label{fig:teaser}
\end{figure}

One major concern of the above solutions lies on the reliability of the adopted tracking algorithms for label generation. The cutting-edge visual trackers are still not robust enough and may easily suffer from drift or other tracking failures under challenging scenarios. However, many existing methods~\cite{trackingnet} directly adopt the tracking results as the generated annotation, leading to unreliable video annotation. For one thing, these approaches mostly fail to select reliable tracking results by measuring their quality. For another, there does not exist an effective mechanism to automatically refine or correct the inaccurate tracking results.  Compared to tracking algorithms based on visual content, temporal interpolation with box geometry modeling across frames are often more robust against severe occlusion and target appearance variations. Some recent attempts~\cite{VI} have also been made to combine visual trackers with temporal interpolation based on heuristics for more accurate bounding box annotation. Nevertheless, how to jointly model appearance and temporal geometry in a principled manner is still an open question in the video annotation community.

Based on the above observation, we propose Video Annotation by Selection-and-Refinement (VASR), a new framework for video annotation with target bounding boxes. Following prior works, we first run an existing tracker initialized by sparse manual annotations to obtain preliminary tracking results (Fig.~\ref{fig:teaser} (b)). Our core idea is to select high-quality tracking results from the preliminary ones and produce reliable annotations through additional bounding box refinement. To this end, we design a temporal assessment network (T-Assess Net) which predicts a quality score (Fig.~\ref{fig:teaser} (b)) for tracking results by modeling their temporal dependencies across frames, providing a criteria for tracking results selection. To correct the potential mistakes of the selected tracking results, we further develop a visual-geometry refinement network (VG-Refine Net), which is able to infer target regions (Fig.~\ref{fig:teaser} (c)) by considering both target appearance and temporal relationship of bounding box geometry.

Both T-Assess Net and VG-Refine Net are learned in a data-driven manner, acting as a principled way to facilitate video annotation. Compared to prior works, our method mainly operates in an offline manner and does not require heavy human interaction. Therefore, we can better focus on improving the accuracy and reliability of the generated annotation at a more flexible complexity budget.

In summary, the contribution of our method is threefold.
\begin{itemize}
    \item We propose a new framework to assist video annotation through bounding box selection and refinement, which not only reduces the human labor but also significantly improves the quality of generated annotations. 
    \item We present new architecture designs to implement the above idea, where the T-Assess Net measures the quality of tracking results through temporal correlation modeling and the VG-Refine Net is able to further improve tracking accuracy by integrating both appearance and temporal geometry cues.   
    \item We empirically show that our method can reduce the amount of manual labels by 94.0\% and that tracking algorithms trained with our generated annotations compares on-par with and even more robust than their counterparts using manual annotations.    
\end{itemize}

\begin{figure}
    \centering
    \includegraphics [width=0.98\linewidth]{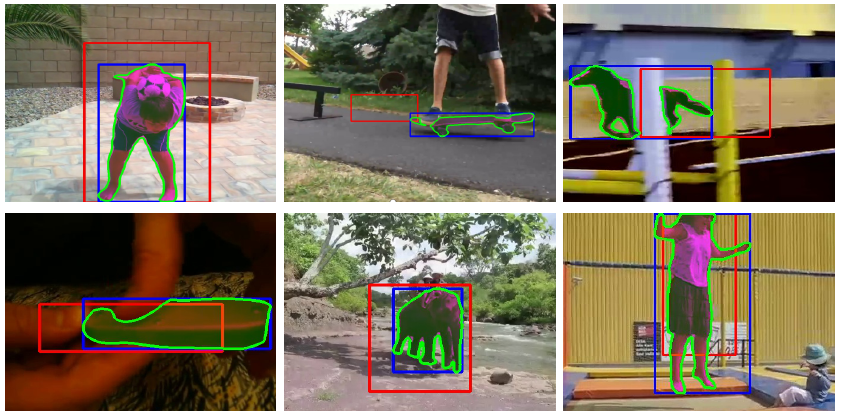}
    \caption{Comparison of TrackingNet\cite{trackingnet} annotations generated using a tracking algorithms\cite{mueller2017context} (Red) and produced by our VASR after selection and refinement. Green contours denotes the target region inferred by the proposed VG-Refine Net.}
    \label{fig:my_label}
\end{figure}

Extensive evaluation results verify that our method can serve as an effective tool to further push the state-of-the-art tracking performance by augmenting training data with high-quality annotations (See Fig.~\ref{fig:my_label}) at a manageable cost. Our project is
available on the website: \url{https://github.com/Daikenan/VASR}.

\begin{figure*}[t]
	\begin{center}
		\includegraphics[width=0.90\linewidth]{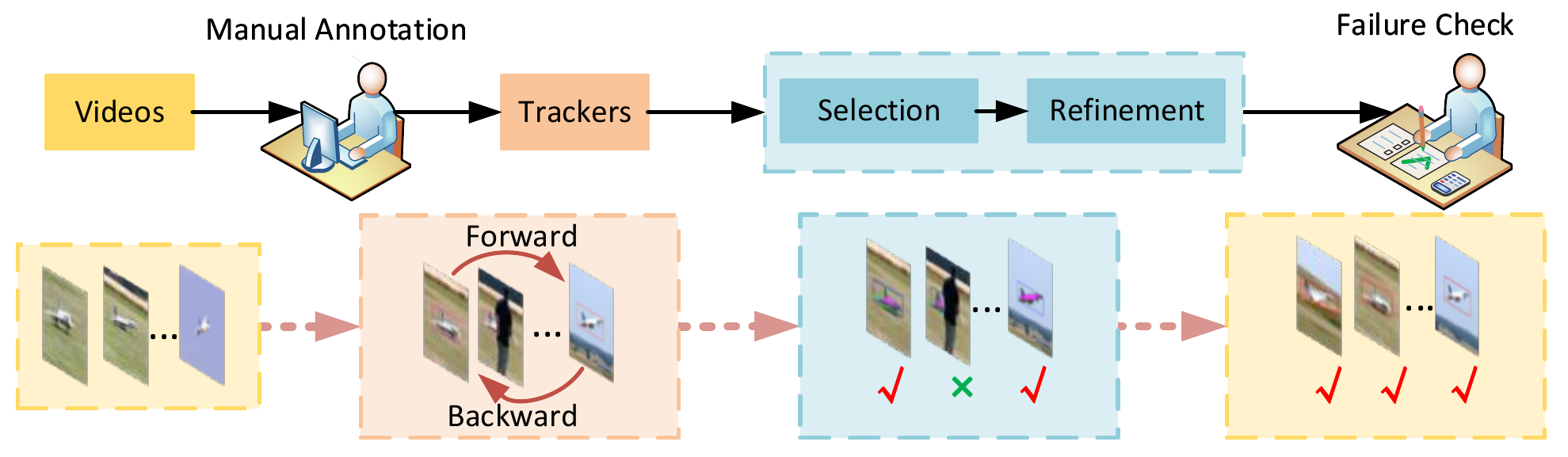}
	\end{center}
	\caption{Pipeline of our VASR method.}
	\label{fig:pipeline}
\end{figure*}

\section{Related Work}
\noindent \textbf{Tracking datasets}.
With the rapid development of the tracking task, many large-scale tracking datasets have appeared, such as LaSOT~\cite{lasot}, TrackingNet~\cite{trackingnet}, GOT-10k~\cite{huang2019got10k} and OxUva~\cite{valmadre2018oxuva}. Among them, LaSOT has 1400 sequences with 70 categories. There are more than 3.5M frames in total where bounding boxes of targets are all annotated manually. GOT-10k is also a purely manually labeled dataset, which contains over 10000 video segments with 1.5M annotations. Although this manual annotating manner can guarantee the quality of labels, it is labor intensive and expensive. To increase the efficiency of labeling, some datasets choose to annotate labels sparsely, such as TrackingNet and OxUva. TrackingNet has more than 30,000 sequences and the total length of the dataset exceeds 14M frames. It labels one bounding box every 30 frames, while other unlabeled frames obtain their labels automatically by an interpolation method where $\rm STAPLE_{CA}$~\cite{mueller2017context} is used for tracking. However, this way will affect the quality of annotations. The labels of the target in the intermediate frames are not precise enough, and the confidence information is lacking. Existing large-scale datasets all have problems that it is difficult to trade-off the efficiency and quality when generating annotations.\par

\noindent \textbf{Single object tracking}. At present, single object tracking develops rapidly and has made a lot of progress, especially for the methods based on deep learning. In terms of whether the model is fine-tuned online, existing trackers can be divided into offline training methods~\cite{SiamFC,SiameseRPN,dasiamrpn,siamrpn++,xu2020siamfc++,guo2020siamcar,chen2020siamban} and online update methods~\cite{MDNet,ATOM,prdimp,dimp}. SiamFC~\cite{SiamFC} proposes a fully convolutional Siamese network, where the cross-correlation layer is used to calculate the similarity between the template and search region. SiamRPN~\cite{SiameseRPN} applies the region proposal network into the Siamese-based tracker and proposes the classification and regression branches, which improves both accuracy and speed. To make the tracker adapt to deep networks and improve performance further, SiamRPN++~\cite{siamrpn++} proposes a sampling strategy to break the spatial invariance restriction. For online update trackers, ATOM~\cite{ATOM} proposes a tracking architecture consisting of dedicated target estimation and classification components. To improve the discriminative ability, DiMP~\cite{dimp} introduces a discriminative learning loss, which significantly improves the tracking performance. These trackers have performed quite well when dealing with short sequences.\par

\noindent\textbf{One-shot learning segmentation}. This task also develops rapidly, including \cite{wang2019siamMask,yan2020alpha}. Given the template in the initial frame, methods need to segment target areas in subsequent frames. \cite{jain2014supervoxel} constructs a spatial-temporal graph from video sequence using supervoxels and optical flow. While \cite{wang2017super} proposes a video object segmentation method based on super-trajectory, which is an efficient video representation and can capture the potential space-temporal structure information. These types of algorithms are often used as a good scale estimator in single object tracking.

\noindent \textbf{Trajectory annotation tasks}. In order to reduce the cost of labor, some methods that generate annotations automatically for large-scale video datasets have been proposed. A common practice is to label few key frames sparsely by annotators, and use linear interpolation to calculate the bounding boxes of other unlabeled frames between key frames, such as VIPER-GT~\cite{mihalcik2003ViPER} and LabelMe~\cite{yuen2009labelme}. These methods cannot handle complex situations, e.g. targets moving nonlinearly. To deal with difficult videos better, VATIC~\cite{vondrick2010efficiently} learns a discriminative classifier which is implemented by a fast linear SVM. It gives high scores on positive bounding boxes and low scores for negatives, where the feature of one bounding box consists of HOG and color histogram features. Besides, \cite{vondrick2011video} implements a constrained tracker and dynamic programming algorithms to determine which frames need to be labeled manually. The problem is cast as active learning to obtain highly accurate tracks. In \cite{manen2017pathtrack}, the manual annotation manner is replaced by path supervision for fast annotation. That is, the annotator collects a path annotation with the cursor, which is approximate and does not provide the scale of the object. Given path annotations and object detections as inputs, PathTrack~\cite{manen2017pathtrack} firstly labels each detection with a provisional trajectory and generates detection clusters. Then in the second step, the most probable trajectory is computed via ST shortest paths for each cluster in a detection linkage step. To further reduce the burden of annotators, ScribbleBox~\cite{chen2020scribblebox} introduces an interactive annotation framework where the annotator does not need to watch the full video, and only inspects the automatically determined key frames. It outputs two types of annotations including tracked boxes and masks inside these tracks. For tracking, a parametric curve with few control points is used to annotate bounding boxes by approximating the trajectory, where the annotator can interactively correct. For segmentation, scribbles are exploited as a form of human input and a scribble propagation network is proposed to correct the segmentation masks.\par

\begin{figure*}[htbp]
	\begin{center}
		\includegraphics[width=0.98\linewidth]{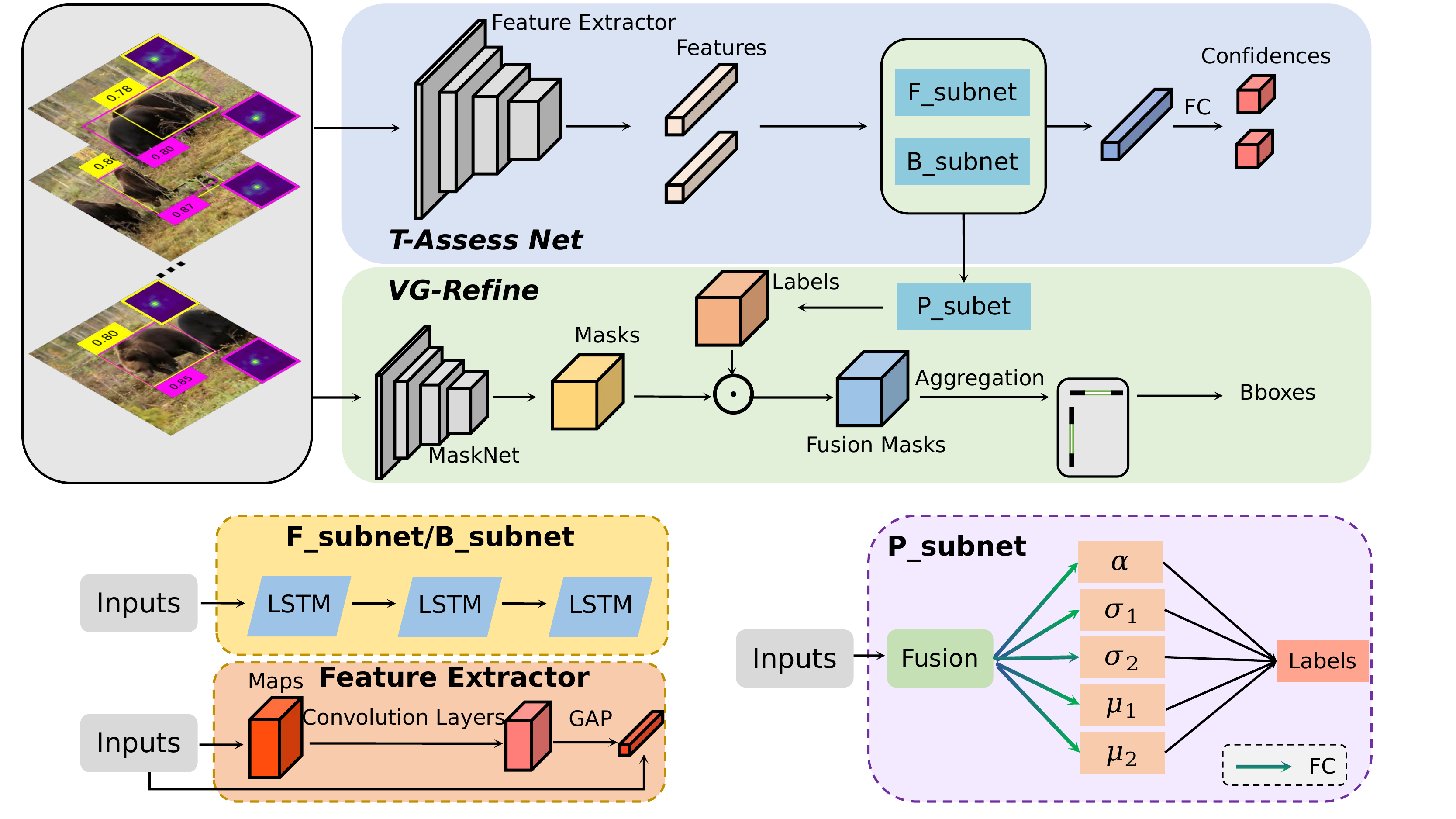}
	\end{center}
	\vspace{-4mm}
	\caption{Architecture of our proposed VASR method.}
	\label{fig-overall_framwork}
	\vspace{-5mm}
\end{figure*}

\section{Annotation with VASR}
At the core of our VASR method is the proposed T-Assess Net and VG-Refine Net which measure the quality of preliminary bounding box labels and perform further label refinement, leading to more accurate automatic labeling.
In the following, we first overview our video annotation framework in Sec~\ref{sec:overview}. In Sec~\ref{sec:architecture}, the detailed architecture designs are provided. Finally, Sec~\ref{sec:train} discusses how to train and apply our approach to achieve high-quality bounding box labels.

\subsection{Overview}\label{sec:overview}
Fig.~\ref{fig:pipeline} overviews the pipeline of the proposed VASR method. Given a video sequence, we first ask human annotators to label a sparse set of frames (\eg, label one frame for every 30 frames). We then adopt an off-the-shelf visual tracker~\cite{dimp} to generate tracking results for each frame as preliminary annotations. To alleviate tracking failures, we split each video at the manually labeled frames into short-term snippets, where the first and last frames of each snippet contain manually labeled bounding boxes. For each snippet, we perform forward and backward tracking use the manual annotation in the first the last frame, respectively, to initialize the tracker which predicts a response map, a target bounding box and its tracking score for each frame. By merging the tracking results of all the snippets, we obtain the forward and backward tracking results for the entire video.

The preliminary tracking results may inevitably contain failure cases. Therefore, we measure the quality of the tracking results and select the more reliable tracking result from forward and backward tracking for each frame. We then perform a bounding box refinement scheme to further improve the quality of the selected tracking results, giving rise to the output annotations. For frames whose forward and backward tracking qualities are both under a predefined threshold, we label them as tracking failures, and resort to additional human annotations. The above process is learned and conducted by the proposed T-Assess Net and VG-Refine Net.

\subsection{Architecture Design}\label{sec:architecture}
\paragraph{T-Assess Net.} 
The input to T-Assess Net contains the initial tracking results $\big\{\bm{b}_i^d, o_i^d, \bm{R}_i^d| i=1,2,\ldots,L, d\in\{\mathcal{F}, \mathcal{B}\}\big\}$ of $L$ consecutive frames, where $\bm{b}_i^d$, $o_i^d$, and $\bm{R}_i^d$ represent the bounding box position, tracker confidence, and response map, respectively, for the $i$-th frame produced by \cite{dimp}, and $d$ indicates whether the result is generated by forward ($d=\mathcal{F}$) or backward ($d=\mathcal{B}$) tracking. The T-Assess Net consists of a feature extractor and a sequential confidence predictor. The feature extractor aims to encode the appearance information of the response map $\bm{R}_i^d$ with a convolutional network, producing a $c$-dimensional feature vector for each input response map. The feature vector is then concatenated with its corresponding bounding box coordinates and tracker confidence, leading to a $c+5$ dimensional compact representation of each tracking result.

The above feature mainly characterizes the spatial, appearance, and confidence information of individual tracking result. To capture the correlation and variation patterns of tracking results in the temporal domain, we design the sequential predictor using three Long Short-Term Memory (LSTM) \cite{lstm} layers with $L$ time steps followed by a fully connected layer. It processes the feature vectors of the $L$ input frames in a sequential manner and predicts a quality score $g_i^d$ for each frame. We use two separate sequential predictors with the same architecture to handle forward and backward tracking results, respectively, which is shown to deliver more superior performance than using a single sequential predictor in our experiments. See Fig.~\ref{fig-overall_framwork} for an illustration of the architecture.

\paragraph{VG-Refine Net.}
The T-Assess Net provides an important cue for selecting high-quality tracking results. To further improve the accuracy of the selected results, we design the VG-Refine Net which learns to perform bounding box refinement by jointly considering both visual and geometric information. To encode visual appearance, we adopt the pretrained MaskNet proposed in \cite{alpha_refine} to predict an initial target segmentation map. Specifically, we crop search regions in the $i$-th frame centered at the two bounding boxes $\bm{b}_i^{\mathcal{F}}$ and $\bm{b}_i^{\mathcal{B}}$ generated by forward and backward tracking, respectively, with twice the size of the bound boxes. Based on the search regions and the initial target template, the MaskNet predicts two initial target segmentation masks $\bm{\tilde{S}}_i^{d} \in \mathbb{R}^{P\times Q}$ corresponding to forward ($d=\mathcal{F}$) and backward ($d=\mathcal{B}$) tracking. 

As shown in our experiments, refinement by considering visual information alone is still not reliable. Therefore, we adopt geometric information to further ensure tracking accuracy. Rather than using a handcrafted geometric interpolation model as in \cite{VI}
we propose a trainable geometric module which can learn to capture the geometric relationships of target locations in the temporal domain. Inspired by the success of T-Assess Net in sequential modeling, we design the geometric module by adopting a similar architecture as T-Asses Net, which also contains the feature extractor and a sequential predictor based on LSTMs. It takes the tracking results of $L$ consecutive frames as input, learns to encode their geometric variations, and predicts a set of Gaussian weight parameters $\bm{\theta}_i^{d}=\{\mu_1, \mu_2, \sigma_1, \sigma_2, \alpha\}$ corresponding to the target segmentation mask $\bm{\tilde{S}}_i^{d}$. We then generate the geometric weight map $\bm{W}_i^{d} \in \mathbb{R}^{P\times Q}$ according to the predicted parameters as follows:
\begin{equation}
\bm{W}_i^{d}(x, y)=\text{exp}\left(-\alpha\left(\frac{\left(x-\mu_{1}\right)^{2}}{\sigma_{1}^{2}}+\frac{\left(y-\mu_{2}\right)^{2}}{\sigma_{2}^{2}}\right)\right),
\end{equation}
where $\bm{W}_i^{d}(x, y)$ denotes the weight value located at coordinate $(x,y)$. The final segmentation mask $\bm{S}_i^{d}$ is achieved by an element-wise multiplication between the initial mask and weight map  $\bm{S}_i^{d} = \bm{\tilde{S}}_i^{d} \odot \bm{W}_i^{d}$. See Fig.~\ref{fig-overall_framwork} for an illustration of the architecture.

\subsection{Training and Inference}\label{sec:train}

\begin{figure}
    \centering
    \label{fig_f}
    \includegraphics [width=0.98\linewidth]{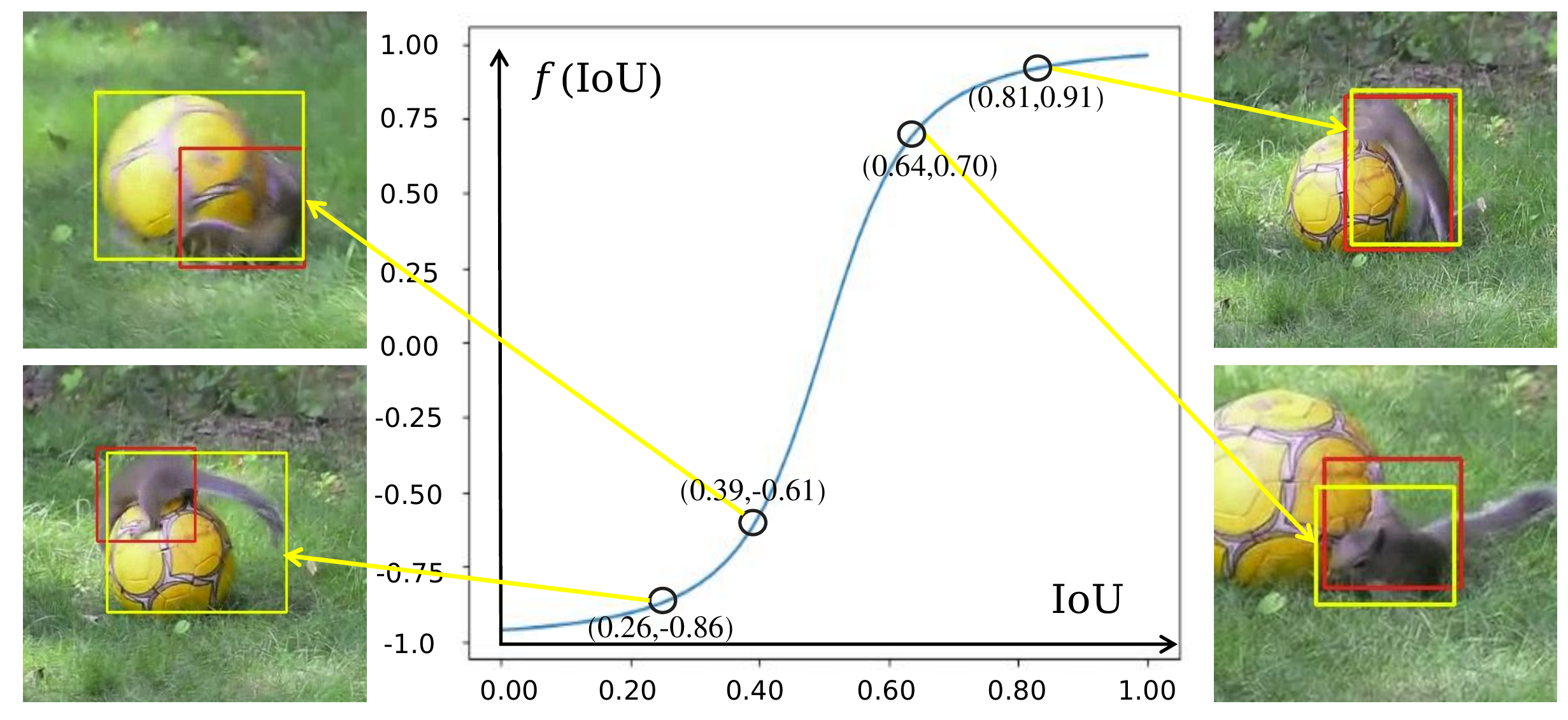}
    \caption{The non-linear function \eqref{eq_fiou} used to compute quality scores. Red and yellow bounding boxes indicate manually annotated ground truth and tracking results, respectively. The quality score can better measure the tracking reliability than IoU.}
    \label{fig:qscore}
\end{figure}

\paragraph{Training.} 
The proposed T-Assess and VG-Refine Net can be learned using video sequences with ground truth annotations.
For each training video, we first split it into video snippets of 30 frames to collect the forward and backward tracking results according to the procedure described in Sec~\ref{sec:overview}.
%
%
We then densely select short-term snippets with a fixed length of 20 frames from all the training videos, which together with the corresponding tracking results serve as input training samples to our method. 

The quality of each tracking result is measured according to its Intersection over Union ($\text{IoU}$) with the ground truth. We empirically find that tracking results with $\text{IoU}>0.5$ are mostly reliable, while $\text{IoU}<0.5$ mainly corresponds to low-quality results. Therefore, we convert the IoU of each tracking result to a quality score $\hat{g}$ using a non-linear function $f(\cdot)$ as follows:

\begin{equation}
\hat{g}=f(\text{IoU})=\frac{\sqrt[\beta]{\alpha}(\text{IoU}-0.5)}{\sqrt[\beta]{1+\alpha(\text{IoU}-0.5)^{\beta}}},
\label{eq_fiou}
\end{equation}
where the hyper parameters $\alpha$ and $\beta$ are empirically set to 50 and 2, respectively. As shown in Fig.~\ref{fig:qscore}, the quality score can effectively measure the reliability of tracking results and is treated as the ground truth of our T-Assess Net. 
The T-Assess Net takes the tracking results of a snippet as input, predicts their quality scores $\left\{g_i^b| i=1,2,\ldots,20; b\in \left\{\mathcal{F}, \mathcal{B} \right\} \right \}$, and is trained by minimizing their differences to the ground truth:
\begin{equation}
    L_{\text{conf}} = \sum_{i} \sum_{b} \|g_i^b - \hat{g}_i^b\|_2^2.
\end{equation}

Although the ground truth bounding box is unable to precisely delineate the target contour, it provides an important cue that each row and column going through the box region also has overlap with the target region. In light of the above observation, we propose to train VG-Refine Net using box-level supervision under a multiple instance learning setting. 
To this end, we first generate a binary box mask $\bm{M}_i$ for each frame according to the ground truth bounding box. The box mask has the same spatial size of $P \times Q$ as the segmentation mask $\bm{S}_i^d$, with $\bm{M}_i(x,y)=1$ indicating the pixel located at $(x,y)$ belonging to the ground truth bounding box regions, and $\bm{M}_i(x,y)=0$ otherwise. Both the predicted segmentation mask and the ground truth box mask can then be aggregated along the vertical and horizontal direction as follows.
\begin{equation}
\begin{split}
\bm{s}_i^{d,h} &= A^h(\bm{S}_i^d),\\
\bm{m}_i^{h} &= A^h(\bm{M}_i^h),
\end{split}
\end{equation}
where $A^h$ denotes the horizontal aggregation operator which map each row of the input mask into a scalar. $\bm{s}_i^{d,h} \in \mathbb{R}^{P}$ and $\bm{m}_i^{h} \in \mathbb{R}^{P}$ denote the aggregated results for segmentation mask and box mask, respectively. The vertically aggregated results $\bm{s}_i^{d,v} \in \mathbb{R}^{Q}$ and $\bm{m}_i^{v} \in \mathbb{R}^{Q}$ can be obtained in a similar manner through aggregation along the vertical direction. The VG-Refine Net can then be trained by minimizing the aggregated results of the predicted segmentation mask and ground truth box mask:
\begin{equation}
    L_{\text{reg}} = \sum_i \sum_d \|\bm{s}_i^{d,v} - \bm{m}_i^{v}\|_2^2 + \|\bm{s}_i^{d,h} - \bm{m}_i^{h}\|_2^2.
\end{equation}

There are many choices for the aggregation operator including one-dimensional max pooling, average pooling, summation, \etc. We design the following rectified accumulation operator which achieves the best performance in our experiments:
\begin{equation}
    A^h(\bm{M}) = \text{max}\left(1, \sum_{x=1}^{Q}\left(\bm{M}(x, \cdot)\right)\right),
\end{equation}
where the summation is independently conducted along each row of the input mask. The vertical aggregation operator $A^v$ is defined in a similar manner by replacing the row summation with the column summation.

It should be noted that a similar multiple instance learning idea has been explored in a concurrent work~\cite{boxinst}. However, \cite{boxinst} adopts max pooling for aggregation and focuses on instance segmentation, while our ultimate goal is to infer an accurate bounding box from the predicted mask rather than a precise target segmentation.

\paragraph{Inference.}
During inference, we feed a snippet of 20 frames and their corresponding forward/backward tracking results into our method. 
The T-Assess and VG-Refine Net predict the quality score $g_i^d$ and the target segmentation mask $\bm{S}_i^d \in \mathbb{R}^{P\times Q}$, respectively, for each tracking result, with frame index $i=1,2,\ldots,20$ and direction indicator $d\in \left\{\mathcal{F}, \mathcal{B}\right\}$. To infer a refined bounding box from the segmentation mask $\bm{S}_i^d$, we first aggregate the predicted mask along the vertical and horizontal directions using the rectified accumulation operator, producing the aggregated results $\bm{s}_i^{d,v} \in \mathbb{R}^{Q}$ and $\bm{s}_i^{d,h} \in \mathbb{R}^{P}$, respectively. We then select two sets of coordinates $\{x|\bm{s}_i^{d,v}(x) > \tau\}$ and $\{y|\bm{s}_i^{d,h}(y) > \tau\}$ according to the aggregated results. The minimum and maximum coordinates of the above two sets forms the corner coordinates of the refined bounding box denoted as $\bm{\tilde{b}}_i^d=(x_{\text{min}}, y_{\text{min}}, x_{\text{max}}, y_{\text{max}})$.
Given the predicted quality scores $g_i^d$ and the refined bounding boxes $\bm{\tilde{b}}_i^d$ for forward and backward tracking at the $i$-th frame, we regard the refined bounding box with higher quality score as the output box annotation if its score is higher than a pre-defined threshold (0), otherwise, we mark the $i$-th frame as a failure frame which requires additional manual annotations.

\section{Experiments}

\subsection{Implementation}\label{sec:impl}
The LaSOT dataset is one of the few large-scale tracking datasets whose ground truth are all manually annotated. Therefore, our proposed video annotation method is trained on the LaSOT training set, and then applied to the training set of both LaSOT and TrackingNet to produce bounding box annotations. To annotate the TrackingNet dataset, we use all the training videos of LaSOT dataset to train our annotation method. To annotate the LaSOT dataset, we adopt a cross-validation manner by first splitting the 1120 training sequences of LaSOT into two subsets\footnote{The LaSOT dataset contains 1120 training sequences belonging to 70 categories with each category containing 16 sequences. We uniformly split the training set into two subsets such that each subset contains 8 sequences of each category.}, and then use the method trained on one subset to annotate the other one until annotations for all the 1120 sequences are generated. We use 3.3\% of all the ground truth as manual annotations to initialize our method. Finally, there are 2.7\% and 1.7\% of video frames in the LaSOT and TrackingNet training set, respectively, being labeled as failure frames by our method, which are further annotated using ground truth. We implement this work using Tensorflow on a PC machine with 8 NVIDIA GTX2080Ti GPU. Data preparation and training for the entire network on LaSOT will take approximately 2 weeks and inference speed is 30 FPS on a single GPU .

To verify the effectiveness of our method, we train 5 state-of-the-art trackers, including SiamRPN++~\cite{siamrpn++}, SiamFC++~\cite{xu2020siamfc++}, ATOM~\cite{ATOM}, DiMP~\cite{dimp} and PrDiMP~\cite{prdimp}, on the training set of LaSOT and TrackingNet using the original and our generated bounding box annotations, respectively. The trained trackers are compared on the test set of LaSOT, TrackingNet, UAV, and GOT10K.


\subsection{Comparison Results}
\begin{figure}
    \centering
    \begin{tabular}{c@{}c@{}c}
      \includegraphics[width=0.5\linewidth]{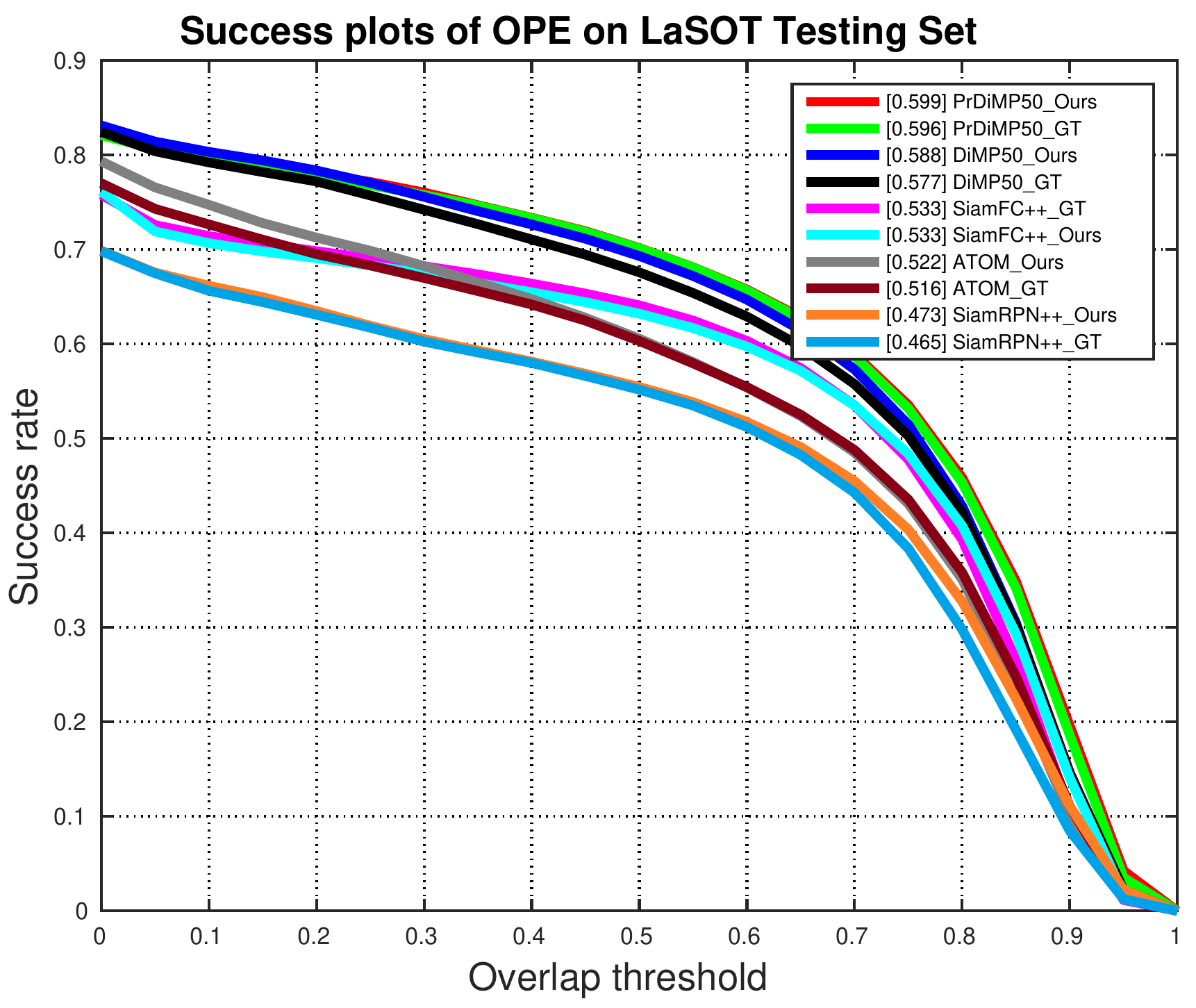} &\

      \includegraphics[width=0.5\linewidth]{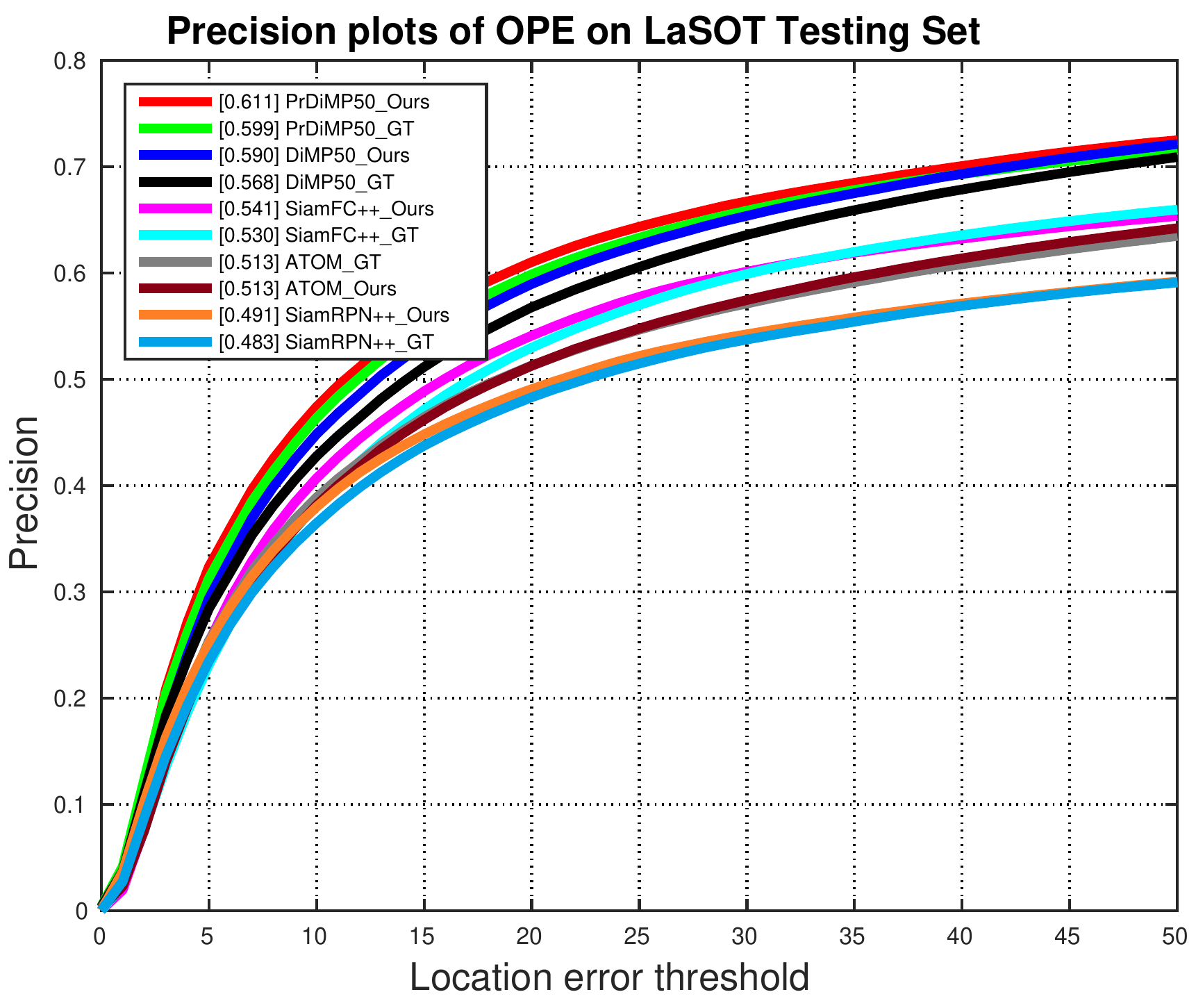} &\\
    \end{tabular}

    \caption{Tracking performance on LaSOT dataset by using our LaSOT annotations (Ours) and manual LaSOT annotations (GT).}
    \label{fig:lasot_plot}
\end{figure}

\begin{figure}
    \centering

    \begin{tabular}{c@{}c@{}c}
      \includegraphics[width=0.5\linewidth]{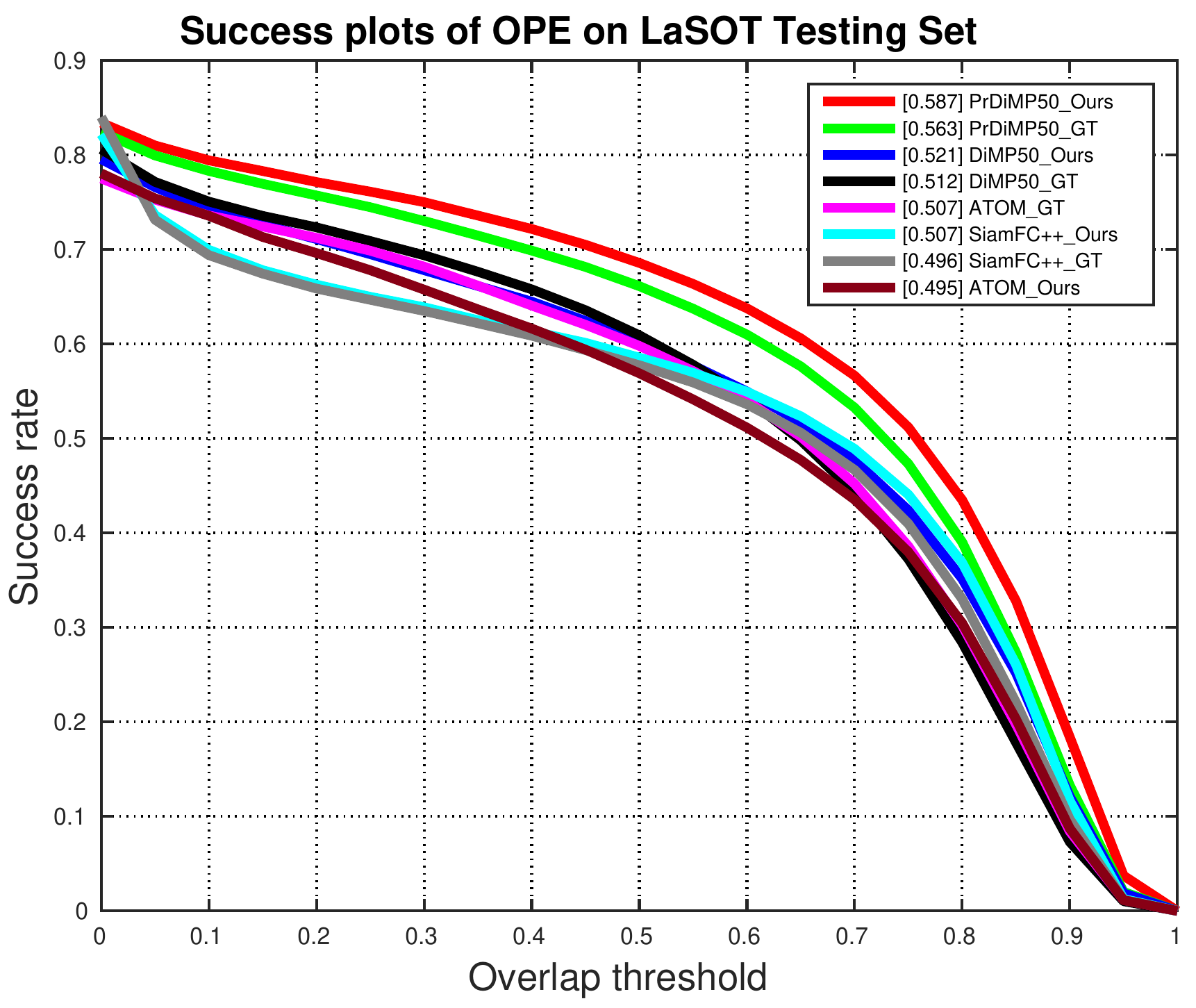} &\

      \includegraphics[width=0.5\linewidth]{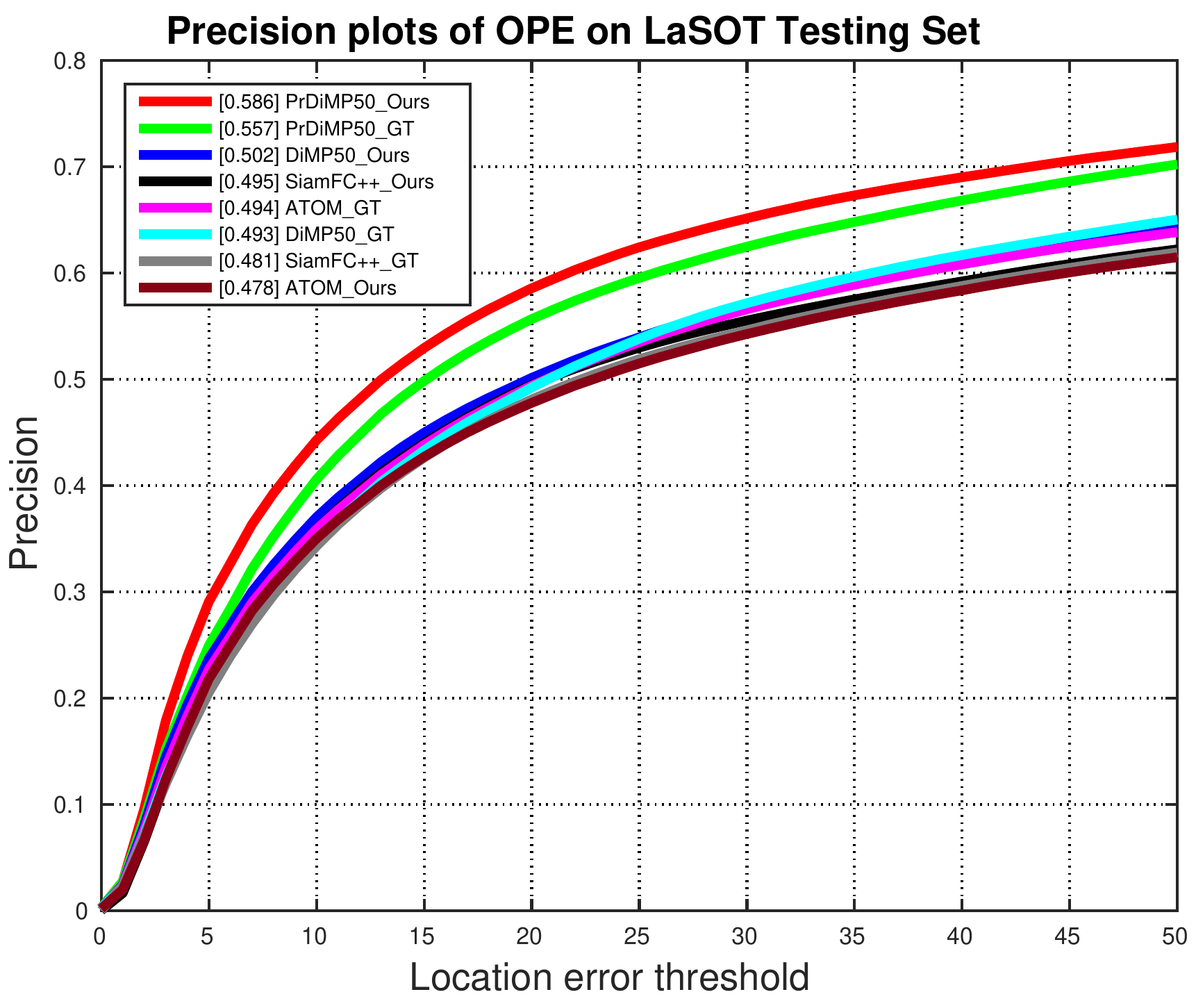} &\\
    \end{tabular}
    \caption{Tracking performance on LaSOT dataset by using our TrackingNet annotations (Ours) and manual TrackingNet annotations (GT).}
    \label{fig:trackingnet_plot}
\end{figure}

Tab.~\ref{tab_lasot_training_performance} and Tab.~\ref{tab_trackingnet_training_performance} reports the comparison results of all the trackers trained on LaSOT and TrackingNet using the original and our generated annotations. From Tab.~\ref{tab_lasot_training_performance}, it can be observed that the compared trackers trained using our generated annotations perform on par with their counterparts trained using the original ground truth on the LaSOT dataset. When training on the TrackingNet datasets (Tab.~\ref{tab_trackingnet_training_performance}), our generated annotations can even yield more superior performance than the original ground truth annotations. The reason might be attributed to the fact that the LaSOT training set are fully annotated by human annotators, while over 96\% of the annotations provided by the TrackingNet train set are produced using tracking algorithms. As shown in Fig.~\ref{fig:my_label}, the evaluation results justify the effectiveness our annotation method and the quality of our generated annotations. 
The results in Tab.~\ref{tab_trackingnet_training_performance} also confirm that our method can well generalize across different datasets.  

\begin{table*}[htbp]

\centering
\caption{Tracking performance on different dataset by using ours LaSOT annotations(Ours) and manual LaSOT annotations(GT).The {\color[HTML]{FE0000} \textbf{red}} results indicate that our annotations achieve the same or better results than the mannual ones.}\label{tab_lasot_training_performance}
\begin{tabular}{cccccccccccc}
\hline
                              &      & \multicolumn{2}{c}{SiamRPN++}                                                 & \multicolumn{2}{c}{SiamFC++}                                                  & \multicolumn{2}{c}{ATOM}                                                      & \multicolumn{2}{c}{DiMP}                                                      & \multicolumn{2}{c}{PrDiMP}                                                    \\
                              &      & Succ                                  & Pre                                   & Succ                                  & Pre                                   & Succ                                  & Pre                                   & Succ                                  & Pre                                   & Succ                                  & Pre                                   \\
                              
                              & GT   & 0.615                                 & 0.594                                 & 0.697                                 & 0.625                                 & 0.704                                 & 0.641                                 & 0.717                                 & 0.650                                 & 0.684                                 & 0.609                                 \\ 
\multirow{-2}{*}{TrackingNet} & Ours & {\color[HTML]{FE0000} \textbf{0.631}} & {\color[HTML]{FE0000} \textbf{0.601}} & {\color[HTML]{FE0000} \textbf{0.698}} & {\color[HTML]{FE0000} \textbf{0.634}} & 0.702                                 & 0.634                                 & 0.715                                 & {\color[HTML]{FE0000} \textbf{0.652}} & {\color[HTML]{FE0000} \textbf{0.688}} & {\color[HTML]{FE0000} \textbf{0.619}} \\ \hline
                              & GT   & 0.552                                 & 0.750                                 & 0.573                                 & 0.769                                 & 0.625                                 & 0.831                                 & 0.629                                 & 0.833                                 & 0.604                                 & 0.793                                 \\
\multirow{-2}{*}{UAV123}      & Ours & {\color[HTML]{FE0000} \textbf{0.557}} & 0.745                                 & {\color[HTML]{FE0000} \textbf{0.577}} & {\color[HTML]{FE0000} \textbf{0.770}} & {\color[HTML]{FE0000} \textbf{0.625}} & {\color[HTML]{FE0000} \textbf{0.842}} & {\color[HTML]{FE0000} \textbf{0.634}} & {\color[HTML]{FE0000} \textbf{0.846}} & 0.598                                 & 0.790                                 \\ \hline
                              &      & AO                                    & $\text{SR}_{0.75}$                              & AO                                    & $\text{SR}_{0.75}$                              & AO                                    & $\text{SR}_{0.75}$                              & AO                                    & $\text{SR}_{0.75}$                              & AO                                    & $\text{SR}_{0.75}$                              \\
                              & GT   & 0.438                                 & 0.230                                 & 0.535                                 & 0.367                                 & 0..562                                & 0.409                                 & 0.593                                 & 0.444                                 & 0.554                                 & 0.416                                 \\
\multirow{-2}{*}{GOT10K}      & Ours & {\color[HTML]{FE0000} \textbf{0.439}} & {\color[HTML]{FE0000} \textbf{0.260}} & {\color[HTML]{FE0000} \textbf{0.549}} & {\color[HTML]{FE0000} \textbf{0.391}} & {\color[HTML]{FE0000} \textbf{0.563}} & {\color[HTML]{FE0000} \textbf{0.416}} & {\color[HTML]{FE0000} \textbf{0.596}} & {\color[HTML]{FE0000} \textbf{0.460}} & {\color[HTML]{FE0000} \textbf{0.570}} & {\color[HTML]{FE0000} \textbf{0.444}} \\ \hline
\end{tabular}
\vspace{-3mm}
\end{table*}

\subsection{Ablation Study}
To have an in-depth understanding of the contributions brought by each component of our method, we perform additional ablation study on the LaSOT dataset. We train different variants of our method and apply them to generate annotations for the training set of LaSOT as described in Sec.~\ref{sec:impl}. We adopt three metrics to measure the accuracy of the generated annotations, including mIoU, Acc@0.5, and Acc@0.7. mIoU denotes the mean IoU of all the generated annotations over ground truth bounding boxes. Acc@threshold indicates the percentage of generated annotations whose IoU are above the threshold.

\noindent \textbf{Impact of Tracking Result Selection. }
Based on the quality scores predicted by our T-Assess Net, our annotation method is able to select the reliable tracking result from forward and backward tracking, and determine whether tracking fails on the current frame. To measure the impact of tracking result selection on the final generated annotations, we compare 4 variants of our method. We denote Fwd and Bwd as two variants which do not perform selection and use the all tracking results from forward and backward tracking, respectively. Sel denotes the variant that selects the more reliable tracking results generated by forward and backward tracking, while Sel-fail adds additional failure detection to Sel. Tab.~\ref{ablation_conf} demonstrates the annotation accuracy by the 4 variants on the LaSOT training set. Sel yields higher accuracy than both Fwd and Bwd, indicating the effectiveness of tracking result selection. 
2.7\% of all the 1120 frames are labeled as tracking failure by Sel-fail, which require additional manual annotation and are not included for accuracy computation. However, the accuracy gain by filtering out the 30 frames is considerable.

\begin{figure}
    \centering
    \includegraphics[width=1.0\linewidth]{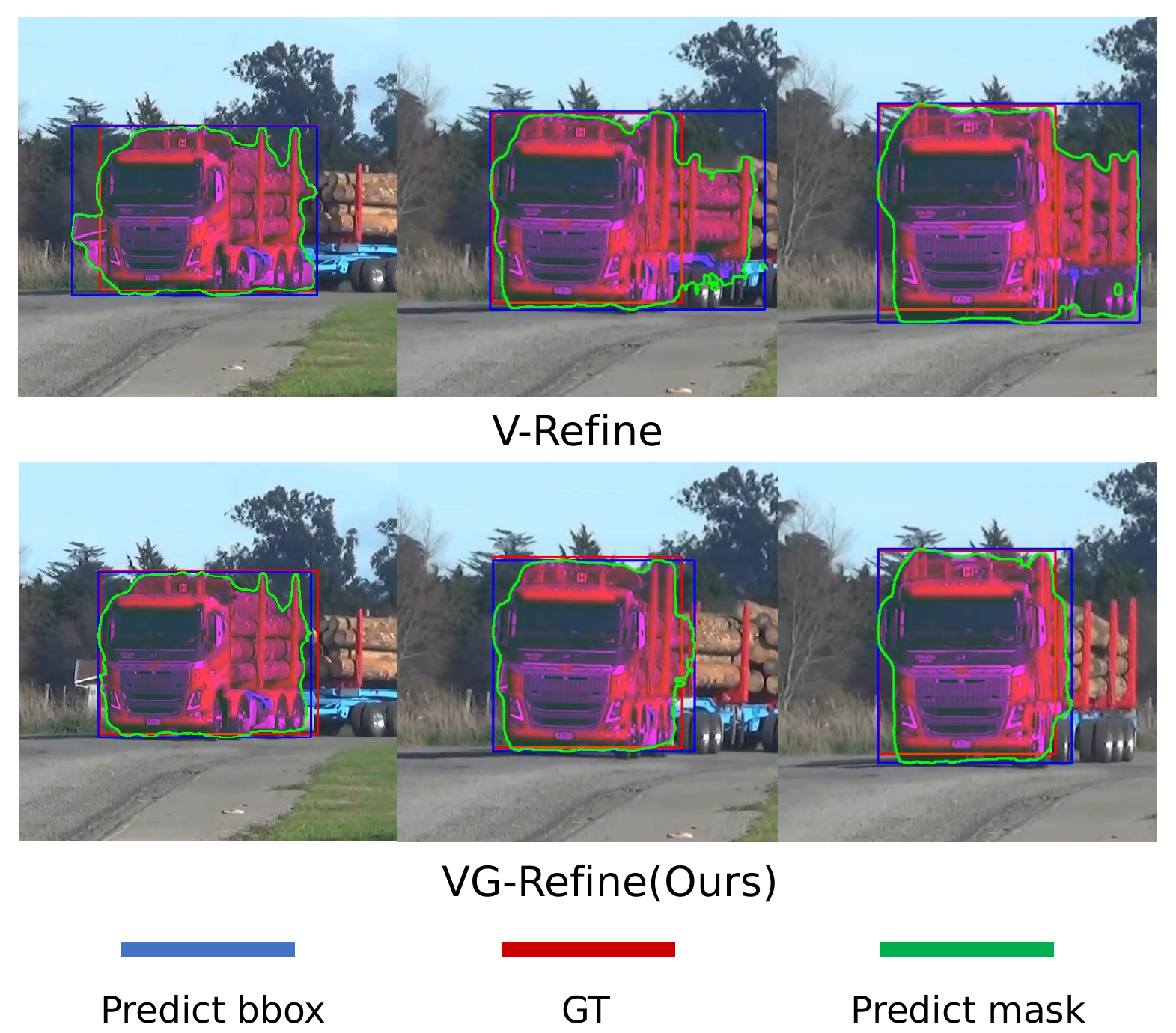}
    \caption{This figure visualizes the comparison between our VG-Refine Net and V-Refine. }
    \label{fig:GW}
\end{figure}

\noindent \textbf{Impact of Tracking Result Refinement. }
Our VG-Refine Net combines target appearance and temporal geometry information through a learning based method to improve the accuracy of the generated annotations. To analyze its impact, we compare 4 variants our method. Among others, w/o-Refine does not perform any refinement and directly using the selected tracking results as the generated annotations. V-Refine performs bounding box refinement based on target region inference considering only the visual appearance information. VI-Refine combines target region inference with geometric interpolation, where geometric interpolation is performed in a handcrafted manner following \cite{VI} rather than a learning based approach. VG-Refine denotes our proposed method. Tab.~\ref{ablation_reg} shows their annotation accuracy on the LaSOT training set. Fig.~\ref{fig:GW} visualizes the comparison between our VG-Refine Net and V-Refine.
By only considering the appearance information, the annotation accuracy of V-Refine is even worse than the original tracking results.
By further enforcing a handcrafted geometric interpolation scheme, VI-Refine can slightly improve the annotation accuracy. In comparison, the proposed VG-Refine integrates target appearance and temporal geometry in a learning based manner, which deliver more superior performance than both V-Refine and VI-Refine. 

To further demonstrate the advantages of our learning based geometry model over handcrafted ones, we further compare our method with \cite{VI} which blends the tracking output with a geometric interpolation result. Tab.~\ref{compare_with_vi} compares the annotation accuracy on the GOT10K datasets, where we use the results reported by \cite{VI} for fair comparison. Our method performs favorably against \cite{VI} in terms of Acc@0.7.

\noindent \textbf{Effectiveness of Temporal Modeling. }
Both T-Select and VG-Refine Net adopts LSTM architectures to model temporal consistency of the tracking results. To verify its effectiveness, we compare our method with its variant that replaces LSTM layers with fully connected ones. As shown in Tab.~\ref{tab_ablation_lstm}, the annotation accuracy is significantly improved by using LSTM layers, suggesting the importance of temporal modeling during video annotation.

\noindent \textbf{Impact of Annotation Amount. }
Due to the high cost of manual annotations, only a few existing large-scale tracking benchmarks~\cite{huang2019got10k,lasot,imagenet} perform exhaustive manual annotations, while others only provide manual annotations for a subset of frames.
To analyze its impact on the tracking performance, 
we collect 3 subsets of the LaSOT training set containing 100\%, 3.33\%, and 1.67\% of all the manual annotations, respectively. \emph{More detailed descriptions can be found in the supplementary material.}

\begin{table}[]
\caption{The effect of results selection.}
\label{ablation_conf}
\centering
\begin{tabular}{c|c|c|c}
\hline
         & mIoU                                  & Acc@0.5                             & Acc@0.7                             \\ \hline
Fwd      & 0.834                                 & 96.3\%                                 & 90.7\%                                 \\ \hline
Bwd      & 0.833                                 & 96.3\%                                 & 90.7\%                                 \\ \hline
Sel      & 0.845                              & 96.6\%                                       & 87.6\%                              \\ \hline
Sel-fail & {\color[HTML]{FE0000} \textbf{0.851}} & {\color[HTML]{FE0000} \textbf{97.0\%}} & {\color[HTML]{FE0000} \textbf{91.1\%}} \\ \hline
\end{tabular}
\vspace{-5mm}
\end{table}

\begin{table}[]
\caption{The effect of results refinement.}
\label{ablation_reg}
\centering
\begin{tabular}{c|c|c|c}
\hline
           & mIoU                                  & Acc@0.5                             & Acc@0.7                             \\ \hline
w/o-Refine & 0.851                                 & 97.0\%                                 & 91.1\%                                 \\ \hline
V-Refine   & 0.845                                 & 96.4\%                                 & 86.9\%                                 \\ \hline
VI-Refine  & 0.853                                 & 97.1\%                                 & 91.1\%                                 \\ \hline
VG-Refine  & {\color[HTML]{FE0000} \textbf{0.865}} & {\color[HTML]{FE0000} \textbf{97.3\%}} & {\color[HTML]{FE0000} \textbf{91.3\%}} \\ \hline
\end{tabular}
\end{table}

\begin{table}[]
\caption{The effect of learning sequential information. }\label{tab_ablation_lstm}
\begin{tabular}{c|c|c|c|c}
\hline
     & Miou                                  & Acc@0.5                        & Acc@0.7                        & Err Rate                               \\ \hline
FC   & 0.859                                 & 96.8\%                                 & 90.8\%                                 & 0.34\%                                 \\ \hline
LSTM & {\color[HTML]{FE0000} \textbf{0.865}} & {\color[HTML]{FE0000} \textbf{97.3\%}} & {\color[HTML]{FE0000} \textbf{91.3\%}} & {\color[HTML]{FE0000} \textbf{0.25\%}} \\ \hline
\end{tabular}
\end{table}


\begin{table}[]

\caption{The table shows our performance on GOT10K validation set compared to VI\cite{VI}.}\label{compare_with_vi}
\begin{tabular}{c|c|c|c|c}
\hline
                         &          & Acc@0.5 & Acc@0.7                           & mIoU \\ \hline
                         & VI{\cite{VI}} & -          & 0.75                                 & -    \\ \cline{2-5} 
\multirow{-2}{*}{GOT10K} & Ours     & 0.96       & {\color[HTML]{FE0000} \textbf{0.90}} & 0.83 \\ \hline
\end{tabular}
\end{table}

\begin{table}[]
\label{tab_trackingnet_training_performance}
\caption{Tracking performance on \textbf{TrackingNet} test set by using ours TrackingNet annotations(Ours) and original TrackingNet annotations(GT) for training.}
\begin{tabular}{ccccccc}
\hline
     & \multicolumn{2}{c}{SiamFC++}                                                  & \multicolumn{2}{c}{DiMP}                                                      & \multicolumn{2}{c}{PrDiMP}                                                    \\
     & Succ                                  & Pre                                   & Succ                                  & Pre                                   & Succ                                  & Pre                                   \\

GT   & 0.754                                 & 0.705                                 & 0.717                                 & 0.661                                 & 0.736                                 & 0.683                                 \\
Ours & {\color[HTML]{FE0000} \textbf{0.770}} & {\color[HTML]{FE0000} \textbf{0.722}} & {\color[HTML]{FE0000} \textbf{0.741}} & {\color[HTML]{FE0000} \textbf{0.682}} & {\color[HTML]{FE0000} \textbf{0.762}} & {\color[HTML]{FE0000} \textbf{0.706}} \\ \hline
\end{tabular}
\end{table}

\begin{table}[]
\caption{Tracking performance on \textbf{GOT10K} test set by using ours TrackingNet annotations(Ours) and original TrackingNet annotations(GT) for training.}\label{tab_trackingnet_training_performance}
\begin{tabular}{ccccccc}
\hline
     & \multicolumn{2}{c}{SiamFC++}                                                  & \multicolumn{2}{c}{DiMP}                                                      & \multicolumn{2}{c}{PrDiMP}                                                    \\
     & AO                                    & $\text{SR}_{0.75}$                              & AO                                    & $\text{SR}_{0.75}$                              & AO                                    & $\text{SR}_{0.75}$                              \\
GT   & 0.533                                 & 0.363                                 & 0.546                                 & 0.349                                 & 0.576                                 & 0.436                                 \\
Ours & {\color[HTML]{FE0000} \textbf{0.569}} & {\color[HTML]{FE0000} \textbf{0.442}} & {\color[HTML]{FE0000} \textbf{0.570}} & {\color[HTML]{FE0000} \textbf{0.441}} & {\color[HTML]{FE0000} \textbf{0.589}} & {\color[HTML]{FE0000} \textbf{0.488}} \\ \hline
\end{tabular}
\vspace{-3mm}
\end{table}

\section{Conclusion}
This paper presents a video annotation method through a selection-and-refinement scheme implemented by a T-Select Net and a VG-Refine Net. The T-Select Net aims select reliable preliminary annotations generated by tracking algorithms by modeling their temporal coherence. The VG-Refine Net integrates both target appearance and temporal geometry through a learning based approach to further improve the annotation accuracy. Experiments on large-scale tracking benchmarks show that our method can effectively reduce the human labors by 94.0\% by delivering high quality video annotations in an automatic manner, which significantly pushes the state-of-the-art tracking performance.     

{\small
\bibliographystyle{ieee_fullname}

\bibliography{ref}

\begin{thebibliography}{10}\itemsep=-1pt

\bibitem{SiamFC}
Luca Bertinetto, Jack Valmadre, Joao~F Henriques, Andrea Vedaldi, and Philip~HS
  Torr.
\newblock Fully-convolutional siamese networks for object tracking.
\newblock In {\em European Conference on Computer Vision}, pages 850--865,
  2016.

\bibitem{dimp}
Goutam Bhat, Martin Danelljan, Luc~Van Gool, and Radu Timofte.
\newblock Learning discriminative model prediction for tracking.
\newblock In {\em IEEE International Conference on Computer Vision}, pages
  6182--6191, 2019.

\bibitem{chen2020scribblebox}
Bowen Chen, Huan Ling, Xiaohui Zeng, Gao Jun, Ziyue Xu, and Sanja Fidler.
\newblock Scribblebox: Interactive annotation framework for video object
  segmentation.
\newblock {\em arXiv preprint arXiv:2008.09721}, 2020.

\bibitem{chen2020siamban}
Zedu Chen, Bineng Zhong, Guorong Li, Shengping Zhang, and Rongrong Ji.
\newblock Siamese box adaptive network for visual tracking.
\newblock In {\em IEEE Conference on Computer Vision and Pattern Recognition},
  pages 6668--6677, 2020.

\bibitem{ATOM}
Martin Danelljan, Goutam Bhat, Fahad~Shahbaz Khan, and Michael Felsberg.
\newblock {Atom}: {A}ccurate tracking by overlap maximization.
\newblock In {\em IEEE Conference on Computer Vision and Pattern Recognition},
  pages 4660--4669, 2019.

\bibitem{prdimp}
Martin Danelljan, Luc~Van Gool, and Radu Timofte.
\newblock Probabilistic regression for visual tracking.
\newblock In {\em IEEE Conference on Computer Vision and Pattern Recognition},
  pages 7183--7192, 2020.

\bibitem{lasot}
Heng Fan, Liting Lin, Fan Yang, Peng Chu, Ge Deng, Sijia Yu, Hexin Bai, Yong
  Xu, Chunyuan Liao, and Haibin Ling.
\newblock {LaSOT}: {A} high-quality benchmark for large-scale single object
  tracking.
\newblock In {\em IEEE Conference on Computer Vision and Pattern Recognition},
  pages 5374--5383, 2019.

\bibitem{lstm}
Alex Graves.
\newblock {\em Supervised Sequence Labelling with Recurrent Neural Networks},
  volume 385 of {\em Studies in Computational Intelligence}.
\newblock Springer, 2012.

\bibitem{guo2020siamcar}
Dongyan Guo, Jun Wang, Ying Cui, Zhenhua Wang, and Shengyong Chen.
\newblock Siamcar: Siamese fully convolutional classification and regression
  for visual tracking.
\newblock In {\em IEEE Conference on Computer Vision and Pattern Recognition},
  pages 6269--6277, 2020.

\bibitem{huang2019got10k}
Lianghua Huang, Xin Zhao, and Kaiqi Huang.
\newblock Got-10k: A large high-diversity benchmark for generic object tracking
  in the wild.
\newblock {\em IEEE Transactions on Pattern Analysis and Machine Intelligence},
  2019.

\bibitem{jain2014supervoxel}
Suyog~Dutt Jain and Kristen Grauman.
\newblock Supervoxel-consistent foreground propagation in video.
\newblock In {\em European Conference on Computer Vision}, pages 656--671.
  Springer, 2014.

\bibitem{VI}
Alina Kuznetsova, A. Talati, Y. Luo, K. Simmons, and Vittorio Ferrari.
\newblock Efficient video annotation with visual interpolation and frame
  selection guidance.
\newblock {\em CoRR}, abs/2012.12554, 2020.

\bibitem{siamrpn++}
Bo Li, Wei Wu, Qiang Wang, Fangyi Zhang, Junliang Xing, and Junjie Yan.
\newblock {SiamRPN++}: {E}volution of siamese visual tracking with very deep
  networks.
\newblock In {\em IEEE Conference on Computer Vision and Pattern Recognition},
  pages 4282--4291, 2019.

\bibitem{SiameseRPN}
Bo Li, Junjie Yan, Wei Wu, Zheng Zhu, and Xiaolin Hu.
\newblock High performance visual tracking with siamese region proposal
  network.
\newblock In {\em IEEE Conference on Computer Vision and Pattern Recognition},
  pages 8971--8980, 2018.

\bibitem{manen2017pathtrack}
Santiago Manen, Michael Gygli, Dengxin Dai, and Luc Van~Gool.
\newblock Pathtrack: Fast trajectory annotation with path supervision.
\newblock In {\em IEEE International Conference on Computer Vision}, pages
  290--299, 2017.

\bibitem{mihalcik2003ViPER}
David Mihalcik and David Doermann.
\newblock The design and implementation of viper.
\newblock {\em University of Maryland}, 21:22, 2003.

\bibitem{mueller2017context}
Matthias Mueller, Neil Smith, and Bernard Ghanem.
\newblock Context-aware correlation filter tracking.
\newblock In {\em IEEE Conference on Computer Vision and Pattern Recognition},
  pages 1396--1404, 2017.

\bibitem{trackingnet}
Matthias Muller, Adel Bibi, Silvio Giancola, Salman Alsubaihi, and Bernard
  Ghanem.
\newblock {Trackingnet}: {A} large-scale dataset and benchmark for object
  tracking in the wild.
\newblock In {\em European Conference on Computer Vision}, pages 300--317,
  2018.

\bibitem{MDNet}
Hyeonseob Nam and Bohyung Han.
\newblock Learning multi-domain convolutional neural networks for visual
  tracking.
\newblock In {\em IEEE Conference on Computer Vision and Pattern Recognition},
  pages 4293--4302, June 2016.

\bibitem{real2017youtube}
Esteban Real, Jonathon Shlens, Stefano Mazzocchi, Xin Pan, and Vincent
  Vanhoucke.
\newblock Youtube-boundingboxes: A large high-precision human-annotated data
  set for object detection in video.
\newblock In {\em IEEE Conference on Computer Vision and Pattern Recognition},
  pages 5296--5305, 2017.

\bibitem{imagenet}
Olga Russakovsky, Jia Deng, Hao Su, Jonathan Krause, Sanjeev Satheesh, Sean Ma,
  Zhiheng Huang, Andrej Karpathy, Aditya Khosla, Michael Bernstein, et~al.
\newblock Imagenet large scale visual recognition challenge.
\newblock {\em International Journal of Computer Vision}, 115(3):211--252,
  2015.

\bibitem{boxinst}
Zhi Tian, Chunhua Shen, Xinlong Wang, and Hao Chen.
\newblock Boxinst: High-performance instance segmentation with box annotations.
\newblock {\em CoRR}, abs/2012.02310, 2020.

\bibitem{valmadre2018oxuva}
Jack Valmadre, Luca Bertinetto, Joao~F Henriques, Ran Tao, Andrea Vedaldi,
  Arnold~WM Smeulders, Philip~HS Torr, and Efstratios Gavves.
\newblock Long-term tracking in the wild: A benchmark.
\newblock In {\em European Conference on Computer Vision}, pages 670--685,
  2018.

\bibitem{vondrick2011video}
Carl Vondrick and Deva Ramanan.
\newblock Video annotation and tracking with active learning.
\newblock {\em Advances in Neural Information Processing Systems}, 24:28--36,
  2011.

\bibitem{vondrick2010efficiently}
Carl Vondrick, Deva Ramanan, and Donald Patterson.
\newblock Efficiently scaling up video annotation with crowdsourced
  marketplaces.
\newblock In {\em European Conference on Computer Vision}, pages 610--623.
  Springer, 2010.

\bibitem{wang2019siamMask}
Qiang Wang, Li Zhang, Luca Bertinetto, Weiming Hu, and Philip~HS Torr.
\newblock Fast online object tracking and segmentation: A unifying approach.
\newblock In {\em IEEE Conference on Computer Vision and Pattern Recognition},
  pages 1328--1338, 2019.

\bibitem{wang2017super}
Wenguan Wang, Jianbing Shen, Jianwen Xie, and Fatih Porikli.
\newblock Super-trajectory for video segmentation.
\newblock In {\em Proceedings of the IEEE International Conference on Computer
  Vision}, pages 1671--1679, 2017.

\bibitem{xu2020siamfc++}
Yinda Xu, Zeyu Wang, Zuoxin Li, Ye Yuan, and Gang Yu.
\newblock Siamfc++: Towards robust and accurate visual tracking with target
  estimation guidelines.
\newblock In {\em AAAI Conference on Artificial Intelligence}, volume~34, pages
  12549--12556, 2020.

\bibitem{alpha_refine}
Bin Yan, Dong Wang, Huchuan Lu, and Xiaoyun Yang.
\newblock Alpha-refine: Boosting tracking performance by precise bounding box
  estimation.
\newblock {\em CoRR}, abs/2007.02024, 2020.

\bibitem{yan2020alpha}
Bin Yan, Xinyu Zhang, Dong Wang, Huchuan Lu, and Xiaoyun Yang.
\newblock Alpha-refine: Boosting tracking performance by precise bounding box
  estimation.
\newblock {\em arXiv preprint arXiv:2012.06815}, 2020.

\bibitem{yuen2009labelme}
Jenny Yuen, Bryan Russell, Ce Liu, and Antonio Torralba.
\newblock Labelme video: Building a video database with human annotations.
\newblock In {\em IEEE International Conference on Computer Vision}, pages
  1451--1458. IEEE, 2009.

\bibitem{StructSiam}
Yunhua Zhang, Lijun Wang, Jinqing Qi, Dong Wang, Mengyang Feng, and Huchuan Lu.
\newblock Structured siamese network for real-time visual tracking.
\newblock In {\em European Conference on Computer Vision}, pages 351--366,
  2018.

\bibitem{Ocean_2020_ECCV}
Jianlong Fu Bing Li Weiming~Hu Zhipeng~Zhang, Houwen~Peng.
\newblock Ocean: Object-aware anchor-free tracking.
\newblock In {\em European Conference on Computer Vision}, August 2020.

\bibitem{dasiamrpn}
Zheng Zhu, Qiang Wang, Bo Li, Wei Wu, Junjie Yan, and Weiming Hu.
\newblock Distractor-aware siamese networks for visual object tracking.
\newblock In {\em European Conference on Computer Vision}, pages 101--117,
  2018.

\end{thebibliography}
}

\end{document}